\documentclass{article}

\usepackage{arxiv}

\usepackage[utf8]{inputenc} 
\usepackage[T1]{fontenc}    
\usepackage{hyperref}       
\usepackage{url}            
\usepackage{booktabs}       
\usepackage{amsfonts}       
\usepackage{nicefrac}       
\usepackage{microtype}      
\usepackage{lipsum}
\usepackage{graphicx}
\usepackage{tabularx}

\title{Challenges in Pre-Training Graph Neural Networks for Context-Based Fake News Detection: An Evaluation of Current Strategies and Resource Limitations\thanks{Preprint accepted at LREC-COLING 2024}}

\author{
 Gregor Donabauer \\
  Information Science\\
  University of Regensburg\\
  Germany \\
  \texttt{gregor.donabauer@ur.de} \\
   \And
 Udo Kruschwitz \\
  Information Science\\
  University of Regensburg\\
  Germany \\
  \texttt{udo.kruschwitz@ur.de} \\
}

\begin{document}
\maketitle
\begin{abstract}
Pre-training of neural networks has recently revolutionized the field of Natural Language Processing (NLP) and has before demonstrated its effectiveness in computer vision. At the same time, advances around the detection of fake news were mainly driven by the context-based paradigm, where different types of signals (e.g. from social media) form graph-like structures that hold contextual information apart from the news article to classify. We propose to merge these two developments by applying pre-training of Graph Neural Networks (GNNs) in the domain of context-based fake news detection. Our experiments provide an evaluation of different pre-training strategies for graph-based misinformation detection and demonstrate that transfer learning does currently not lead to significant improvements over training a model from scratch in the domain. We argue that a major current issue is the lack of suitable large-scale resources that can be used for pre-training.
\end{abstract}


\section{Introduction}
Fake news spreading online nowadays poses one of the biggest threats to democracy \cite{Shu_2023}, presenting a urgent requirement for the development of robust detection techniques \cite{zhou_zafarani_2020}. Advances in recent years were mainly driven by context-based methods, e.g. \cite{wei-etal-2022-uncertainty,min_et_al_2022,mehta_et_al_2022}. Compared to the content-based paradigm, that aims to identify fake articles based on their textual content, context-based methods include various other signals as additional information \cite{Guo22Survey}. This for example comprises information from social media platforms, like posts associated with news articles, sharing patterns, and the profiles of connected users which are helpful cues for the task at hand \cite{shu_et_al_2019,sheikh-ali-etal-2022-detecting}. These different signals are often put together in graph data structures and can then be aggregated using Graph Neural Networks (GNNs), e.g. \cite{lu_li_2020,rode-hasinger-etal-2022-true}.
At the same time, pre-training of Large Language Models (LLMs) on massive text corpora has recently transformed the field of Natural Language Processing (NLP) as this approach equips the trained models with an understanding of general aspects of written language, e.g. \cite{devlin-etal-2019-bert,gpt3,gpt4}. These advances have led to impressive progress across the entire spectrum of NLP, also in the area of fake news detection, e.g. \cite{raza_2021,hartl-kruschwitz-2022-applying}.
In a study conducted by \cite{hu2020pretraining}, they presented a comprehensive evaluation of pre-training techniques within the context of GNNs. 
Inspired by this progress, we propose to adopt similar strategies to learn universal attributes of heterogeneous social media context graphs relevant for the domain of context-based fake news detection.

We present a set of experiments across two fake news detection datasets evaluating how different pre-training strategies perform in the domain. Towards the end we discuss current limitations of the approach and summarize our observations and findings. In the spirit of LREC-COLING we share our implementations and pre-trained models with the community to foster reproducibility\footnote{\url{https://github.com/doGregor/pretrain_gnns_FakeNewsNet}}.

\section{Related Work}

\subsection{Graph-Based Fake News Detection}

We start by providing a brief contextualization within existing related work on the context-based paradigm of graph-based fake news detection. Sometimes context refers to graph-like conversation patterns of posts related to news on social media platforms \cite{ma-gao-2020-debunking,li-etal-2020-exploiting,wei-etal-2022-uncertainty}. However, many approaches also include user \cite{shantanu_et_al_2020,min_et_al_2022,rode-hasinger-etal-2022-true} or publisher \cite{mehta_goldwasser_2021,mehta_et_al_2022} relationships as contextual information.
Recently, it has been demonstrated that taking into account such information within heterogeneous structures can lead to further improvements \cite{nguyen_et_al_2020,het_transformer_2022,ecir2023}. 
In conclusion, graph structures, particularly heterogeneous ones, offer high potential in modelling diverse dimensions of contextual information resulting in notable improvements in the domain of fake news detection.

\subsection{Pre-Training of Graph Neural Networks}

While pre-training neural networks has become standard in NLP, we will briefly summarize relevant work in graph machine learning below.
The most comprehensive evaluation of pre-training strategies was performed by \cite{hu2020pretraining} in the domain of Biology and Chemistry. Generally, GNN pre-training is mainly employed in specific domains like recommender systems \cite{hao_et_al_2021} and biomedical networks \cite{long_et_al_2022}.
Some concentrate on large-scale heterogeneous graphs, using methods like contrastive learning-based node and schema-level pre-training \cite{jiang_et_al_2021}, node masking \cite{fang_et_al_2022}, or structural clustering with attention aggregation \cite{yang2022self}. Others target large homogeneous graphs with node and graph-level pre-training tasks \cite{Lu_Jiang_Fang_Shi_2021}, or generative approaches applicable to both homogeneous and heterogeneous large-scale graphs \cite{hu_et_al_2020_gptgnn}.
In this study, we aim to bridge the gap of applying GNN pre-training approaches within the domain of graph-based fake news detection.

\section{Methodology}

\subsection{Data and Graph Setup}
\label{data_and_graph_setup}

Existing approaches to context-based fake news detection are primarily associated with only a small number of datasets. This includes Twitter 15 and Twitter 16 \cite{ma-etal-2017-detect}, PHEME \cite{kochkina-etal-2018-one} and WEIBO \cite{weibo_dataset}, which are all very limited in terms of size and social context provided. Apart from that there is FakeNewsNet \cite{fakenewsnet} which stands out as the largest dataset in the domain and offers a comparatively large amount of contextual features. This dataset comprises news articles that have been fact-checked by reputable fact-checking organizations and were found circulating on social media platforms. Furthermore, it includes identifiers for the associated social media data, allowing to retrieve and analyze this information through a recrawling processes. Recent efforts to introduce even more extensive and diverse resources were made by \cite{fb_dataset_2022}, who provide social context cues similar to FakeNewsNet but in greater amount and even with multilingual contents.
However, the vast majority of these datasets heavily depends on contextual information from access-restricted social media platforms, predominantly Twitter. Recently, new restrictions posed on the Twitter API\footnote{\url{https://developer.twitter.com/en/docs/twitter-api}} are making it exceedingly challenging for researchers to gain access to these datasets at a reasonable cost or with a reasonable amount of effort.
Due to these limitations, we conduct our experiments using the FakeNewsNet dataset. A useful characteristic of FakeNewsNet is that it consists of two subsets: \textit{Politifact} (483 graphs) and \textit{Gossipcop} (12,214 graphs). This allows us to conduct pre-training on one subset and perform fine-tuning on the other subset. However, the amount of data that potentially can be used for pre-training of GNNs is very limited compared to the amounts of data that are used for pre-training of LLMs in NLP.
The majority of methods that incorporate social media context through graph structures, primarily target the identification of individual nodes within large graphs. These nodes hold contextual information alongside the news articles to classify. This is different to the approach of \cite{hu2020pretraining}, who focus their methodology on a large number of small graphs, each representing an individual data point. We thus adopt our previous approach \cite{ecir2023} to construct the social media context graphs as we formulated the problem as a graph classification task which makes the setup similar to \cite{hu2020pretraining}. Each graph consists of three different sets of nodes: (1) the actual news article to classify at the center of the graph, (2) tweets/retweets/timeline tweets related to the article, and (3) user profiles of people linked with the aforementioned posts. Each node set is disconnected from the other sets of nodes and all initial node representations are BERT text embeddings. These node sets are linked via multiple types of edges ([\textit{tweet, cites, article}], [\textit{user, posts, tweet}], [\textit{user, posts, retweet}], [\textit{retweet, cites, tweet}] and [\textit{user, posts, timeline tweet}]) and each edge type can be interpreted as an individual adjacency matrix that connects two distinct sets of nodes.
Data setup and hyperparameters in our work are adopted from our previously provided implementations \cite{ecir2023}\footnote{\url{https://github.com/doGregor/Graph-FakeNewsNet}}.

However, one issue we need to solve is that these are heterogeneous graphs, compared to homogeneous graphs used by \cite{hu2020pretraining}. We therefore modify the pre-training strategies to be applicable to graphs with different node/edge types.

\subsection{Pre-Training Objectives}

The pre-training objectives we adopt have been proven most effective in \cite{hu2020pretraining} and are twofold: we use node-level objectives and graph-level objectives. The latter can predominantly be viewed as strategies aimed to learn holistic graph representations useful for the fake/real news graph classification task. On the other hand, the node-level tasks serve as regularization of the GNNs to mitigate the risk of overfitting to the self-supervised graph-level pre-training.

\subsubsection{Node-Level Objectives} We use two different node-level pre-training objectives, namely \textit{node masking} and \textit{context prediction}.

\textbf{\textit{Node masking}} is similar to word masking in pre-training of LLMs. We mask a proportion of 15\% of all nodes with placeholder values at random. Given that the initial node attributes are continuous (BERT embeddings) we simply set all masked nodes to a uniform value of $1.0$ in the dimension of the original features. Since we have different node types within our heterogeneous social media graphs, we only mask nodes of type tweet/retweet/timeline tweet which have demonstrated the biggest impact on classification performance. 
During pre-training we jointly train a GNN that serves as node encoder and a fully connected neural network for feature reconstruction.

\begin{figure}[!h]
\begin{center}
\includegraphics[width=0.5\textwidth]{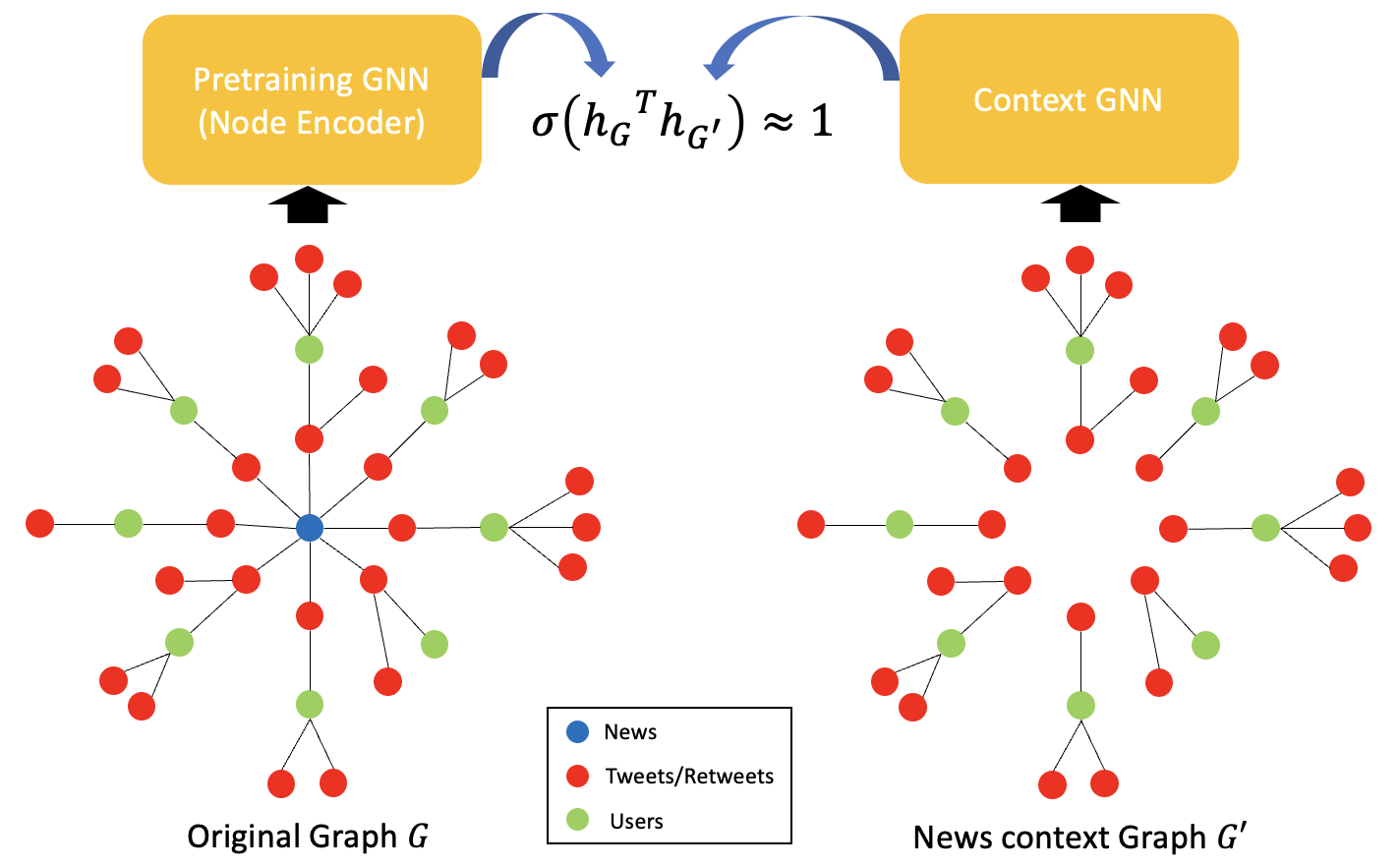} 
\caption{Formulation of pre-training on context prediction between article graph and context graph.}
\label{fig.1}
\end{center}
\end{figure}

For \textbf{\textit{context prediction}} we first create subgraphs of the original graphs. As can be seen in Figure \ref{fig.1} we remove the centre node (news node) from the original graphs to create the subgraphs. We perform negative sampling with a ratio of $1$ to create matching and non-matching pairs of original graphs and subgraphs. In particular, the learning objective is a binary classification task of whether context and actual graph belong to the same news node. During pre-training we jointly train one GNN that serves as node encoder of the original graph and another GNN that serves as node encoder of the context graph. Afterwards we perform mean pooling over all tweet and user nodes and predict whether the embeddings originate from the same graphs. As we are classifying the whole (sub-) graphs this is slightly different to \cite{hu2020pretraining}.

\subsubsection{Graph-Level Objective} As described previously, the graph-level objective serves the purpose of learning holistic graph characteristics that are useful for the upcoming fake/real news graph classification task. We compared the influence of different graph properties on the fake news detection task which revealed that the number of retweets associated with an article seems to have a comparably high influence. We therefore employ the self-supervised task of predicting the number of retweet nodes each graph holds. Since they are part of the same node set as normal tweet nodes and timeline tweet nodes, the GNN faces a challenge in determining the count of retweet nodes solely from the graph properties. Specifically, it cannot read out the number of retweet nodes from a dedicated node set, as these nodes are intertwined with the other tweet nodes (as described in Section \ref{data_and_graph_setup}). The GNN must extract this understanding by leveraging the structural information inherent in the graph, which from our point of view makes this prediction task more challenging.

\subsection{Experiments}

Inside our node encoder we use Heterogeneous Graph Transformer (HGT) GNN layers \cite{hgt_conv} that are based on the Transformer architecture \cite{transformers} which has proven suitable for pre-training in the domain of NLP \cite{devlin-etal-2019-bert} and in other studies on pre-training of GNNs \cite{fang_et_al_2022,hu_et_al_2020_gptgnn}. The encoder consists of two stacked HGT layers where each has 64 input dimensions. The GNN learns distinct sets of weights for each edge type present within the heterogeneous graph and applies them for message passing and aggregation of each node’s neighborhood to update the initial node representations. We first apply the two stacked HGT layers, and finally a fully connected layer that maps the learnt node representations to the target space (graph labels during fine-tuning or node features/context during pre-training).

During pre-training we use a batch-size of 128 and run 50 epochs for node level pre-training tasks with a learning rate of $0.001$; 25 epochs (Politifact) and 50 epochs (Gossipcop) respectively for self-supervised graph-level pre-training, again with a learning rate of $0.001$. For fine-tuning we again use a batch-size of 128 and a learning rate which amounts to $0.001$ (same value as used during pre-training in \cite{hu2020pretraining}).

Our implementations are based on PyTorch Geometric \cite{fey_lenssen_2019}. All experiments are executed using a single NVIDIA RTX A6000 GPU with 48 GB graphical memory size.

We evaluate how different combinations of node-level and graph-level pre-training tasks affect fake news detection performance after fine-tuning. Pre-training is always performed on one full subset (Politifact or Gossipcop), while fine-tuning is performed on 80\% of the left-out subset and tested on the remaining 20\% (with 5-fold cross validation).

Furthermore, we are conducting experiments involving fine-tuning with limited resources. This is particularly interesting due to the high cost associated with acquiring labeled data. This additional evaluation focuses on the utilization of models pre-trained on the Gossipcop dataset as we have a greater amount of pre-training data. Subsequently, we perform fine-tuning using 50 data points ($\sim$10\%) from the Politifact dataset. The model's performance is then evaluated on the complete, remaining test set, using 5-fold cross-validation.

\begin{table*}[h!]
\begin{center}
\begin{tabularx}{\textwidth}{|X|X|r|r|r|r|r|r|r|r|}

    \hline 
    \multicolumn{2}{|c}{Pre-training task} & \multicolumn{4}{|c|}{Fine-tuning on POL} & \multicolumn{4}{c|}{Fine-tuning on GOS} \\
    \hline
    Node-level & Graph-level & P & R & ACC & F1 & P & R & ACC & F1 \\
    \hline

    -- & -- & 0.977 & 0.978 & 0.977 & 0.977 & 0.966 & \textbf{0.979} & \textbf{0.984} & 0.972 \\
    ContextPred & -- & 0.975 & 0.976 & 0.975 & 0.975 & 0.962 & 0.977 & 0.982 & 0.969 \\
    NodeMasking & -- & \textbf{0.981} & \textbf{0.982} & \textbf{0.981} & \textbf{0.981} & 0.959 & 0.978 & 0.981 & 0.968 \\

    \hline

    -- & \#RT & 0.973 & 0.974 & 0.973 & 0.973 & 0.963 & 0.977 & 0.982 & 0.970 \\    
    ContextPred & \#RT & 0.955 & 0.956 & 0.955 & 0.954 & 0.951 & 0.977 & 0.978 & 0.963 \\
    NodeMasking  & \#RT & 0.973 & 0.974 & 0.973 & 0.973 & \textbf{0.969} & 0.978 & \textbf{0.984} & \textbf{0.973} \\
    
    \hline
    
\end{tabularx}
\caption{Precision, Recall, Accuracy and macro F1-scores for fake news detection on the \textit{Politifact} (POL) and \textit{Gossipcop} (GOS) datasets respectively after fine-tuning.}
\label{table:results}
\end{center}
\end{table*}

\begin{table*}[h!]
\begin{center}
\begin{tabularx}{0.5\columnwidth}{|X|X|r|r|}

    \hline
    \multicolumn{2}{|c}{Pre-training task} & \multicolumn{2}{|c|}{Fine-tuning POL} \\
    \hline
    Node-level & Graph-level & ACC & F1 \\
    \hline
    -- & -- & 0.942 & 0.942 \\
    ContextPred & -- & 0.953 & 0.952 \\
    NodeMask. & -- & \textbf{0.965} & \textbf{0.965} \\
    \hline
     -- & \#RT & 0.957 & 0.957 \\
    ContextPred & \#RT & 0.957 & 0.956 \\
    NodeMask. & \#RT & 0.953 & 0.952 \\
    \hline
      
\end{tabularx}
\caption{Accuracy and macro F1 scores for fine-tuning with only 50 labeled samples from Politifact on models pre-trained on the Gossipcop dataset.}
\label{table:results_low_resource}
\end{center}
\end{table*}

\section{Results}

In line with common practice in NLP, we report precision, recall, accuracy and macro F1 scores for all setups \cite[Ch. 4]{Jurafsky24Speech}. All values are the average results obtained by 5-fold cross validation.

As illustrated in Table \ref{table:results} there are only minimal fluctuations between all the setups without any significant differences (using pairwise t-tests at $p<0.01$ and Bonferroni correction on the F1- and Accuracy scores derived from cross-validation across all experimental setups). The most positive impact can be seen for the pre-training strategies involving the node masking approach. For comparison, a strong state-of-the-art baseline excluding any graph-based approaches, achieves $0.956$ ACC and $0.952$ F1-score for Politifact as well as $0.963$ ACC and $0.923$ F1-score for Gossipcop \cite{hartl-kruschwitz-2022-applying}. The best-performing baseline using a graph-based approach achieves $0.975$ ACC and $0.975$ F1-score for Politifact as well as $0.990$ ACC and $0.986$ F1-score for Gossipcop \cite{het_transformer_2022}. Thus, our results on Politifact are marginally (though not significantly) better than the best-reported results in the literature.

However, the potential positive impact on model performance becomes more obvious when fine-tuning on a smaller fraction of the actual labeled data. In that case the learnt properties during pre-training lead to higher accuracy and F1 scores (compare table \ref{table:results_low_resource}) for all setups. Again, these differences are not significant.

\section{Discussion and Limitations}

While our experiments do not demonstrate significant improvements by deploying a pre-training approach, we would like to discuss this in a broader context.

\textbf{Datasets:} The most relevant issue relates to the resources available to conduct our experiments. As already argued in Section \ref{data_and_graph_setup}, their sizes are notably small in comparison to the extensive amount of training data employed for pre-training in the domain of NLP. A general advantage of self-supervised pre-training is that these strategies do not require any labeled data. At the same time, online social media interactions that can serve as context are available in rich amount which means that there is a theoretical solution to this problem. However, current limitations related to the platforms' APIs are making it almost impossible to solve this problem at a reasonable cost or with a reasonable amount of effort. A potential workaround could be data augmentation by, for example, generating graphs used for pre-training at various temporal snapshots during the evolution of user interactions with a news article \cite{song_et_al_2022}. Concluding, the community working on these topics would benefit from solving the problem of limitations in access to shareable resources, similar to the practice in the area of linguistics \cite{cieri-etal-2022-reflections}.

Another factor influencing the overall results could be that performance on FakeNewsNet is already very high, which is why choosing datasets that are more challenging and thus leave more room for improvements could better underscore the potential improvements achievable through pre-training. However, we note that our choice of FakeNewsNet for our exploratory experiments was deliberate, driven by multiple considerations: (1) its status as the largest dataset in the domain, a critical factor for effective pre-training; (2) its richness in social context cues compared to other datasets, allowing to construct heterogeneous graphs with diverse node and edge types; and (3) its division into two subsets, allowing a structured approach of pre-training on one subset followed by fine-tuning on the other subset. Examples for datasets that could be included into follow-up experiments as additional resources are the work by \cite{mehta-goldwasser-2023-interactively} (which is based on \cite{baly-etal-2018-predicting} and introduces richer contextual cues) as well as the ones discussed previously such as Twitter 15 and Twitter 16 \cite{ma-etal-2017-detect}, PHEME \cite{kochkina-etal-2018-one}, WEIBO \cite{weibo_dataset} and FbMultiLingMisinfo \cite{fb_dataset_2022}.

\textbf{Pre-training Objectives:} As shown in Table \ref{table:results}, graph-level pre-training as well as context prediction do not lead to improvements while these have worked well in other domains \cite{hu2020pretraining}. This possibly is due to the tasks' poor alignment with the problem. For context prediction this issue becomes evident from only small fluctuations in loss and accuracy throughout the pre-training phase. One possible explanation is that, unlike in the cases of Biology and Chemistry, where node features hold high expressiveness, our situation places greater emphasis on the graph structure. This was demonstrated in our prior research, where we systematically compared the contributions of node features and graph structure to classification performance \cite{ecir2023}. However, the ego-graph structure of our social media graphs where all context converges onto a single central node is not directly comparable to molecules or proteins (which often exhibit multiple paths between nodes) and thus seems to be less fitting for this pre-training objective's requirements. Moreover, the graph-level task evaluated might generally be less suitable for the problem of fake news detection as we could not observe any positive effects on classification performance. To address the challenges just described, one potential solution lies in employing larger, more densely connected social media context graphs. This is a common setup when framing fake news detection as a node classification task (deviating from the graph classification task alignment we maintained with \cite{hu2020pretraining}).

Another suggestion is to explore approaches similar to \cite{hu_et_al_2020_gptgnn} who apply generative techniques to pre-train GNNs. This could even go beyond their proposed idea of learning to reconstruct the input attributed graph but move towards synthetic data generation, effectively implementing data augmentation. This direction could also serve as a potential strategy to mitigate the problem of data scarcity described previously.

\textbf{Model Size:} While BERT's hidden layers consist of 768 nodes \cite{devlin-etal-2019-bert} and \cite{hu2020pretraining} used GNNs with 300 dimensional layers for pre-training, we experimented with a comparatively shallow model dimension of 64. This aligns with our goal of exploring the general effectiveness of GNN pre-training in the domain. However, using larger networks with higher capacity will potentially lead to better results.

\section{Conclusion}

We demonstrate how a current lack of large-scale resources leads to limitations in pre-training GNNs in the domain of context-based fake news detection. However, there is much potential left as we could observe improvements in experimental setups with small amounts of labelled data. Taking into account all of our observations, we proposed a range of potential avenues for future research in the field.

\section{Ethical Considerations}

We recognize the importance of responsible and ethical research practices, especially in the context of fake news detection.
We provide clear documentation of our methodologies, including the pre-training strategies employed, as well as datasets and evaluation metrics used. We encourage open and transparent discussions about our findings and welcome scrutiny from the research community which is why we also share all our implementations and pre-trained models. Our research aims to contribute to the advancement of AI in a manner that benefits society, while also being aware of potential risks, challenges and limitations. We acknowledge the potential biases in data, algorithms, and models and have made efforts to mitigate these biases to the best of our knowledge.
We acknowledge that fake news detection is a complex field with ethical implications. While our study focuses on technical aspects, we are aware of the broader societal impact.

\section{Acknowledgements}
We would like to thank the anonymous reviewers for their constructive feedback which has helped us improve the paper.

This work was supported by the project COURAGE: A Social Media Companion Safeguarding and Educating Students funded by the Volkswagen Foundation, grant number 95564.

\bibliographystyle{unsrt}  
\bibliography{references}  

\end{document}